\documentclass{article}

\PassOptionsToPackage{numbers}{natbib}

\usepackage[preprint]{nips_2018}

\usepackage[utf8]{inputenc} 
\usepackage[T1]{fontenc}    
\usepackage{hyperref}       
\usepackage{url}            
\usepackage{booktabs}       
\usepackage{amsfonts}       
\usepackage{nicefrac}       
\usepackage{microtype}      
\usepackage{longtable}
\usepackage{graphicx}
\usepackage{subcaption}
\usepackage{authblk}

\usepackage{floatrow}
\newfloatcommand{capbtabbox}{table}[][\FBwidth]

\title{Deep feature transfer between localization and segmentation tasks}

\author[1]{Szu-Yeu Hu}
\author[1]{Andrew Beers}
\author[1]{Ken Chang}
\author[1]{Kathi Höbel}
\author[2]{J. Peter Campbell}
\author[3]{Deniz Erdoguus}
\author[3]{Stratis Ioannidis}
\author[3]{Jennifer Dy}
\author[2,4]{Michael F. Chiang}
\author[1]{Jayashree Kalpathy-Cramer}
\author[1]{James M. Brown}

\affil[1]{Athinoula A. Martinos Center for Biomedical Imaging,  Charlestown, Massachusetts}
\affil[2]{Department of Ophthalmology, Casey Eye Institute, Oregon Health and Science University}
\affil[3]{Department of Electrical and Computer Engineering, Northeastern University, Boston, Massachusetts}
\affil[4]{Department of Medical Informatics and Clinical Epidemiology, Oregon Health and Science University, Portland}
\affil[ ]{\textit {\ Szu-Yeu\char`_Hu@hms.harvard.edu}}

\begin{document}

\maketitle

\begin{abstract}
In this paper, we propose a new pre-training scheme for U-net based image segmentation. We first train the encoding arm as a localization network to predict the center of the target, before extending it into a U-net architecture for segmentation. We apply our proposed method to the problem of segmenting the optic disc from fundus photographs. Our work shows that the features learned by encoding arm can be transferred to the segmentation network to reduce the annotation burden. We propose that an approach could have broad utility for medical image segmentation, and alleviate the burden of delineating complex structures by pre-training on annotations that are much easier to acquire.
\end{abstract}

\section{Introduction}
Deep convolutional neural networks (CNNs) are considered the state-of-the-art for image segmentation problems \cite{lecun1998gradient,guo2018review,lecun2015deep}. 
One limitation of deep CNNs is that they requires large volumes of training data in order to generalize well \cite{chen2014big}. Unfortunately, acquisition of manual segmentation labels is a time-consuming process that requires domain expertise to produce an acceptable ground truth. This is most notable in the domain of medical images, for which accurate delineation of structures require many years of experience on the part of the annotator. 
In contrast, spatial localization of the same structure of interest is often relatively simple, exhibits lower variance, and takes significantly less time to perform.

In this work, we explore the potential of CNNs to learn transferable features from easily-labeled data to tasks that would normally require larger quantities of expertly-labeled data to achieve the same performance. To demonstrate our approach, we consider the task of segmenting the optic disc from retinal fundus photographs. The optic disc is readily discernible from fundus photographs and can have differing coloration and morphology as a consequence of both normal variation and pathology such as glaucoma \cite{joshi2011optic}. Segmentation of the optic disk allows for such characteristics to be readily quantified from conventional fundus images.

\section{Materials and Methods}
\begin{figure}
\includegraphics[width = 9cm]{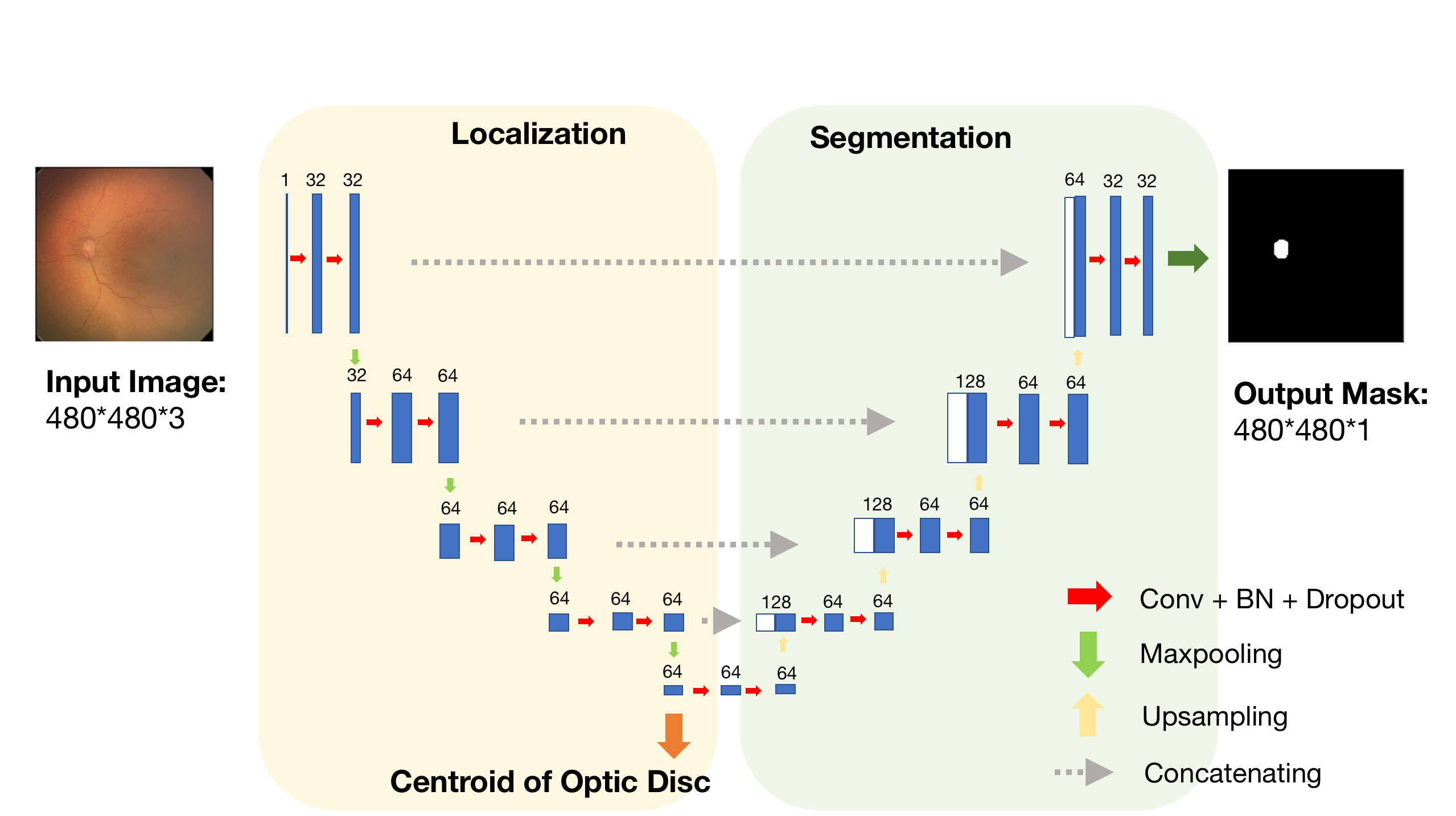}
\centering
\caption{Diagram of the U-net architecture and two phase training scheme for optic localization and segmentation.}
\label{fig:diagram}
\end{figure}

Our method consists of a two phase training scheme of the widely used U-net \cite{ronneberger2015u} architecture, as shown in Figure \ref{fig:diagram}. In the first phase of training, the encoding arm of the U-net is trained independently as an optic disc localization network. After convergence, the network's weights are frozen and the decoding arm is trained as an optic disc segmentation network, re-using the features learned during localization training. We posit that the semantic information learned during training of the encoding arm are transferable to the segmentation task, such as the shape, texture and boundary characteristics of the disc.

\subsection{Data and preprocessing}
Training, validation, and test data sets were created from a database of more than 10,000 de-identified retinal images obtained using a commercially available camera (RetCam; Natus Medical Incorporated) as part of the [details of institution redacted].

We used 9047 images with optic disc center location annotations for the localization task, and a further 92 images labeled with optic disc binary masks for the segmentation task. Images were pre-processed with grayscale conversion, normalization, contrast limited adaptive histogram equalization and gamma correction \cite{brown2018automated}.

\subsection{Optic disc localization}
We formulate the localization task as a regression problem using the encoding arm of a U-net in Figure \ref{fig:diagram}. Pre-processed fundus photographs are provided as input to the network, which undergoes a series of convolutional and pooling operations to produce a volume of image features. These features are passed to a fully connected network without activation (linear) to produce a tuple of ($x, y$) coordinates. We use $3\times3$ convolutions, $2\times2$ max-pooling, batch-normalization, and ReLU activations throughout the network. We also utilize dropout to mitigate overfitting. This network is trained on pre-processed images and their annotated centroid coordinates, minimizing a mean squared error (MSE) cost function using the RMSprop optimizer \citep{Tieleman2012}. We evaluate the performance of the localizer network by calculating the Euclidean distance between the ground truth and predicted centroids. 

\subsection{Optic disc segmentation}
Following training as a localizer, the fully connected layers are removed and the remaining network weights are frozen. The decoding arm is then added to the network, and residual connections added to produce a conventional U-net architecture. Image features learned by the localizer (encoding) network are concatenated with those features learned by the segmentation (decoding) network. The final two layers of the network are a 1x1 convolution followed by a sigmoid activation. The network is trained on the pre-processed images and binary masks representing the optic disc, minimizing a negative log soft Dice loss function using the Adam optimizer \cite{kingma2014adam}. We evaluate the performance of the segmentation network by measuring the Dice overlap between the ground truth and predicted optic disc masks at a probability threshold of 0.5.

\section{Results}
At the first stage of the training (optic disc localization) the MSE values were 132.16 and 206.15 for the validation and test sets, respectively. For optic disc segmentation, the results were evaluated using five-fold cross-validation, and the mean Dice coefficient was 0.88 with 0.1 standard deviation.

\subsection{Comparison with the conventional U-net}
We evaluated the performance of our proposed method with the conventional U-net approach. We trained the full network with the identical layers, hyper-parameters and training data, but without the pre-trained encoder weights (random initialization). The average Dice coefficient was 0.84, with 0.2 standard deviation. The results showed that by learning the features in localization task first, we  improved upon the performance of optic disc segmentation.

\subsection{Performance with fewer training samples}
\begin{figure}
\centering
\begin{subfigure}[t]{0.6\linewidth}
\caption{Dice coefficients for both training schemes}
\includegraphics[width=\linewidth]{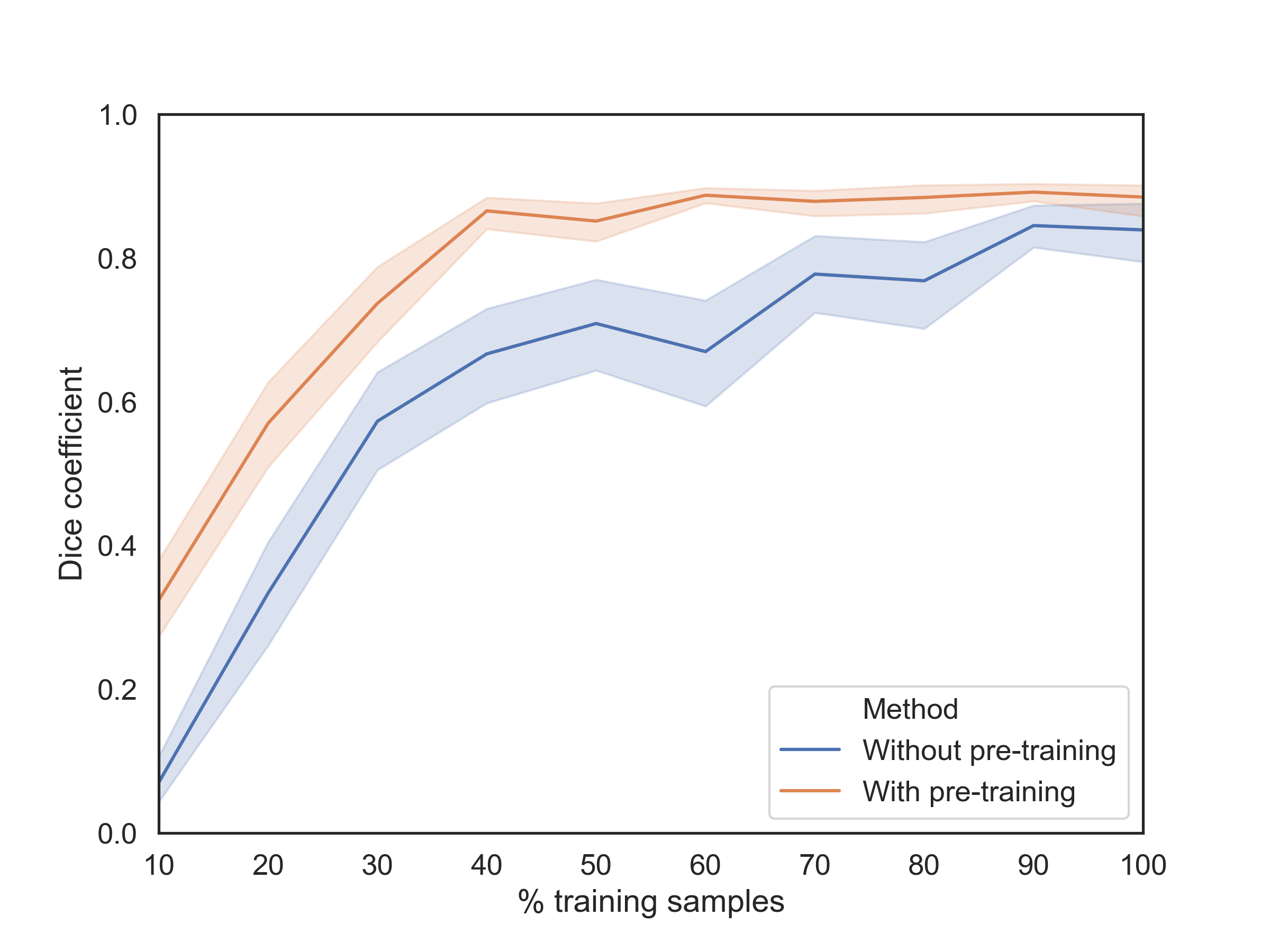}
\label{dice_plot}
\end{subfigure}
\hfill
\begin{subfigure}[t]{0.35\linewidth}
    \centering
    \caption{\emph{p}-values, based on paired $t$-test }
    \vspace{25px}
    \begin{tabular}{cc} \hline
        $\%$ examples & P-value  \\ \hline
        100 & $1.13 \times 10^{-2}$  \\ 
        90 & $3.50 \times 10^{-3}$  \\
        80 & $2.37 \times 10^{-4}$  \\
        70 & $1.51 \times 10^{-3}$  \\
        60 & $2.50 \times 10^{-8}$  \\
        50 & $2.00 \times 10^{-5}$  \\
        40 & $4.95 \times 10^{-8}$  \\
        30 & $1.39 \times 10^{-4}$  \\
        20 & $1.75 \times 10^{-6}$  \\
        10 & $3.97 \times 10^{-12}$  \\
        \hline
    \end{tabular}
    \label{pvalue_table}
\end{subfigure}%
\caption{Dice coefficients of the two training schemes with different proportions of ground truth segmentation data (over all validation splits). The blue line represents the model without the pre-trained encoder, and the orange line is with pre-training as a localizer. The P-values were derived from a paired $t$-test under the null hypothesis that the two models have the same dice coefficients.}
\label{myfigure}

\end{figure}

To see whether the features learned by the localization network can alleviate the demand for manual segmentations, we trained the two models described above (with and without the pre-trained encoder) again, but with different numbers of training samples. Figure \ref{myfigure} shows the Dice coefficient of the models with increasing numbers of training samples. As expected, the Dice coefficient increased in both approaches when using a larger number of images. However, our proposed method affords greater robustness and the differences are most significant when the number of samples is very low. Figure \ref{pvalue_table} shows the results were significantly different at all the sample sizes using a paired $t$-test. The results are consistent with our hypothesis that with the pre-trained encoder, we need fewer segmentation labels to achieve comparable performance.

\section{Conclusion}
In this paper, we introduced a training scheme for optic disc segmentation based on pre-training the encoding arm of a U-net architecture. The approach reduces the demand for expertly-labeled data to achieve good segmentation performance on held out test sets. Though only tested on limited dataset with a relatively simple problem, we propose that this concept is extensible to other imaging modalities and segmentation tasks. We intend to assess the scalability of such an approach on more complex tasks such as brain tumor segmentation, and compare our approach with other pre-training schemes.

\bibliographystyle{unsrtnat}
\bibliography{bib}

\end{document}